\documentclass[journal=jacsat,manuscript=article]{achemso}
\SectionNumbersOn

\usepackage[version=3]{mhchem} % Formula subscripts using \ce{}
\usepackage{graphicx}% Include figure files
\usepackage{dcolumn}% Align table columns on decimal point
\usepackage{bm}% bold math
\usepackage[utf8]{inputenc}
\usepackage[T1]{fontenc}
\usepackage{mathptmx}
\usepackage{etoolbox}
\usepackage{url}
\usepackage{xcolor}
\usepackage{booktabs}
\usepackage{bm}
\usepackage{hyperref}
\usepackage{algorithm}
\usepackage{algpseudocode}
\usepackage{multirow}
\usepackage{makecell}
\usepackage{graphicx}
\usepackage{dcolumn}
\usepackage{bm}
\usepackage{enumitem}
\setlist{nolistsep}
\usepackage[capitalise]{cleveref}

\author{Brijesh FNU}
\affiliation{Department of Mechanical and Materials Engineering, School of Engineering, The University of Alabama at Birmingham (UAB)}
\altaffiliation{These authors contributed equally to this work.}

\author{Viet Thanh Duy Nguyen}
\affiliation{Department of Computer Science, College of Arts and Sciences, The University of Alabama at Birmingham (UAB)}
\altaffiliation{These authors contributed equally to this work.}

\author{Ashima Sharma}
\affiliation{Department of Mechanical and Materials Engineering, School of Engineering, The University of Alabama at Birmingham (UAB)}

\author{Md Harun Or Rashid Molla}
\affiliation{Department of Mechanical and Materials Engineering, School of Engineering, The University of Alabama at Birmingham (UAB)}

\author{Chengyi Xu}
\affiliation{Department of Mechanical and Materials Engineering, School of Engineering, The University of Alabama at Birmingham (UAB)}
\email{cxu@uab.edu}

\author{Truong-Son Hy}
\affiliation{Department of Computer Science, College of Arts and Sciences, The University of Alabama at Birmingham (UAB)}
\email{thy@uab.edu}

\title{Multimodal Machine Learning for Soft High-$k$ Elastomers under Data Scarcity}

\abbreviations{IR, NMR, UV}
\begin{document}

\begin{abstract}
Dielectric materials are critical building blocks for modern electronics such as sensors, actuators, and transistors. With rapid advances in soft and stretchable electronics for emerging human- and robot-interfacing applications, there is a growing need for high-performance dielectric elastomers. However, developing soft elastomers that simultaneously exhibit high dielectric constants ($k$) and low Young's moduli ($E$) remains a major challenge. Although individual elastomer designs have been reported, structured datasets that systematically integrate molecular sequence, dielectric, and mechanical properties are largely unavailable. To address this gap, we curate a compact, high-quality dataset of acrylate-based dielectric elastomers by aggregating experimental results from the past decade. Building on this dataset, we propose a multimodal learning framework leveraging large-scale pretrained polymer representations. These pretrained embeddings transfer chemical and structural knowledge from vast polymer corpora, enabling accurate few-shot prediction of dielectric and mechanical properties and accelerating data-efficient discovery of soft high-$k$ dielectric elastomers. Our data and implementation are publicly available at: \url{https://github.com/HySonLab/Polymers}.
\end{abstract}

Soft and stretchable electronics, including wearable sensors and artificial actuators, demand dielectric elastomers that simultaneously exhibit high dielectric constant ($k$) and low Young’s modulus ($E$). However, achieving this combination remains a major challenge, as inorganic dielectrics offer high permittivity but poor flexibility, while organic polymers provide compliance at the expense of dielectric performance. Designing materials that reconcile these competing properties requires careful molecular engineering. Machine learning (ML) offers a promising route to accelerate such design by uncovering structure–property relationships \cite{amamoto2022data, doan2020machine}. Yet, its effectiveness depends on the availability of structured, high-quality datasets. For soft dielectric elastomers, dielectric and mechanical measurements are typically reported separately across individual studies, and no unified, machine-readable dataset jointly organizes molecular sequence, $k$, and $E$.

To enable data-driven modeling of soft dielectric elastomers, we curated a compact dataset of acrylate-based formulations from peer-reviewed publications over the past decade \cite{Feng2024NatCommun,Yin2024NatCommun54278,Han2023HybridPrestrainLocked,Shi2022ProcessablePHDE,Zhao2022AdvancedAcrylateDE,Yin2021SoftToughFastPolyacrylate,Xiong2023TailoringCrosslinkingNetworks,Ha2008HighElectromechanicalIPN,Liu2022TernaryDipolarDE,Tan2019PolarCrosslinkingDEA,Shao2019AEgOANI,Gao2021ImprovingDielectricAE_PS_RGO,Han2023HybridFabrication,Zhao2018Remarkable,CEJ2021UltrasoftPentablock,Sung2025HighkZwitterionic,Ankit2020HighkUltrastretchable,Li2024SSRNPolarCyano,Wang2023PILFillerDEA,Adeli2024BottlebrushDEG,Park2024GlassTransitionSelfHeal,Zhang2025DEASoftRoboticsReview,Liu2022TernaryDipolarGroups,Shi2018DielectricGels,Bezsudnov2023DEAMaterialsDesign}. Studies reporting both dielectric constant ($k$) and Young’s modulus ($E$) were systematically screened, and only samples with complete and explicitly stated measurements were retained. For each elastomer, the reported chemical composition was mapped to a repeat-unit structure and converted into a standardized SMILES representation. All property values were harmonized to consistent units, with dielectric constants restricted to comparable frequency ranges and Young’s modulus converted to MPa. Records containing ambiguous, incomplete, or non-numeric values were excluded, and duplicate reports were consolidated after removing clear outliers. Each entry retains a direct reference to its original source to ensure traceability and reproducibility.

\begin{figure}[h]
    \centering
    \includegraphics[width=0.85\linewidth]{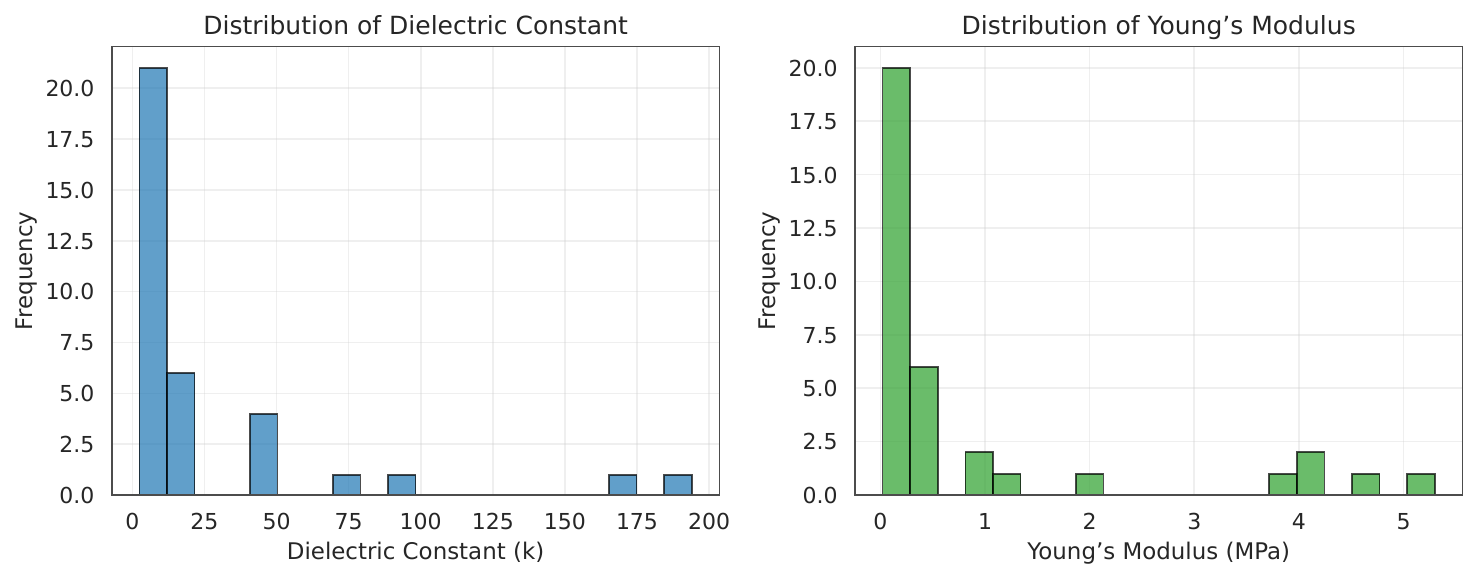}
    \caption{Distributions of dielectric constant ($k$) and Young’s modulus ($E$) across all curated acrylate-based dielectric elastomers.}
    \label{fig:dataset_distributions}
\end{figure}

The final dataset comprises 35 fully standardized elastomer samples. As shown in Figure~\ref{fig:dataset_distributions}, the dielectric constant exhibits a right-skewed distribution, with approximately 71\% of samples falling below $k < 20$ and only a small number of high-$k$ outliers exceeding 100. Young’s modulus values are similarly concentrated in the low-modulus regime, with the majority of samples below 1 MPa, reflecting the predominance of ultra-soft elastomers reported in the literature. These distributional characteristics highlight the intrinsic imbalance of currently available experimental data and motivate the need for data-efficient learning strategies.

\begin{figure}[H]
    \centering
    \includegraphics[width=\linewidth]{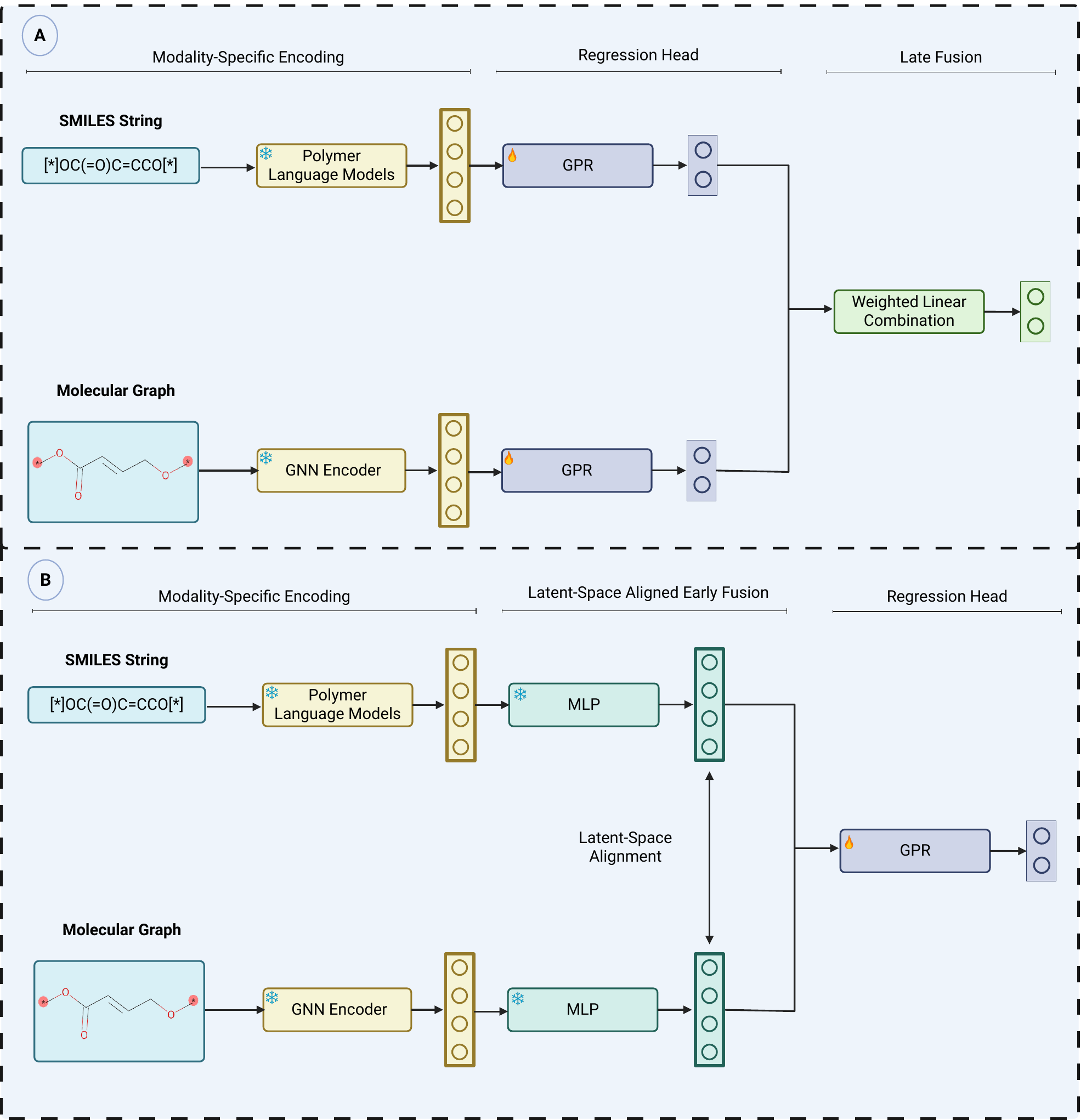}
    \caption{
    \textbf{Overview of the proposed multimodal framework for elastomer property prediction.}
    \textbf{(A) Late fusion:} Pretrained sequence and graph encoders generate modality-specific embeddings, each processed by a Gaussian Process Regressor (GPR); final predictions are combined via weighted averaging.
    \textbf{(B) Latent-aligned early fusion:} Modality-specific embeddings are projected into a shared latent space using lightweight MLP heads trained for cross-modal alignment, fused, and passed to a shared GPR to jointly predict dielectric constant and Young’s modulus.
    }
    \label{fig:framework}
\end{figure}

To enable data-efficient prediction under extreme data scarcity, we develop a multimodal learning framework that integrates pretrained sequence- and graph-based polymer representations (Figure~\ref{fig:framework}). For the sequence modality, polymer SMILES strings are encoded using pretrained Transformer-based polymer language models (e.g., PolyBERT \cite{kuenneth2023polybert} and TransPolymer \cite{xu2023transpolymer}), and fixed-length embeddings are obtained via mean pooling. For the structural modality, polymers are represented as molecular graphs and encoded using a Graph Isomorphism Network (GIN) that we pretrain from scratch in a self-supervised manner on the PI1M polymer database \cite{ma2020pi1m}. The pretraining does not require dielectric or mechanical property labels; instead, masked-atom and bond-type prediction objectives are used to learn transferable chemical representations before downstream adaptation to property prediction. To integrate the two modalities, we evaluate both prediction-level (late) fusion and representation-level (early) fusion. In the latter, each modality-specific embedding is first projected through a lightweight MLP head into a shared latent space and trained using a CLIP-style contrastive objective \cite{radford2021learning}, which encourages aligned representations of the same polymer across modalities before fusion. For downstream regression, we employ a multi-output Gaussian Process Regressor (GPR), which is well-suited for small datasets and enables robust prediction of dielectric constant and Young’s modulus without additional deep parameterization.

All experiments are conducted under an extreme data-scarcity setting using leave-one-out cross-validation (LOOCV) over the curated elastomers. Within each LOOCV iteration, pretrained sequence and graph encoders are kept frozen, and their embeddings are processed through feature standardization, principal component analysis (PCA), and a multi-output Gaussian Process Regressor (GPR). To ensure fair comparison across unimodal and multimodal models, an identical PCA candidate grid is used for all methods. The number of PCA components and GPR hyperparameters are selected via grid search performed exclusively on the training portion of each fold, thereby preventing any information leakage from the held-out sample. The optimized model is then evaluated on the left-out elastomer. Performance is assessed using $R^2$ and RMSE, reported separately for dielectric constant ($k$) and Young’s modulus ($E$), and averaged across both targets. Statistical significance between models is further evaluated using paired tests across LOOCV folds.

We conduct two complementary experiments to evaluate the effectiveness of multimodal integration under extreme data scarcity. The first experiment investigates whether multimodal integration provides benefits beyond unimodal representations. In this setting, each modality, sequence-based (Morgan fingerprints, PolyBERT, TransPolymer) and graph-based (pretrained GIN), is evaluated independently within the same regression framework to quantify the predictive capacity of each representation. For the multimodal configuration, we integrate the strongest-performing encoders from each modality, namely TransPolymer for sequence representations and the pretrained GIN for graph representations, to ensure a fair and performance-driven comparison. The second experiment examines how different fusion strategies affect multimodal performance. Specifically, we compare naive early fusion (concatenation or averaging), prediction-level late fusion, and latent-space aligned early fusion. This design isolates whether explicit cross-modal alignment is necessary for effective integration in the low-data regime.

\begin{figure}[ht]
\centering
\includegraphics[width=\linewidth]{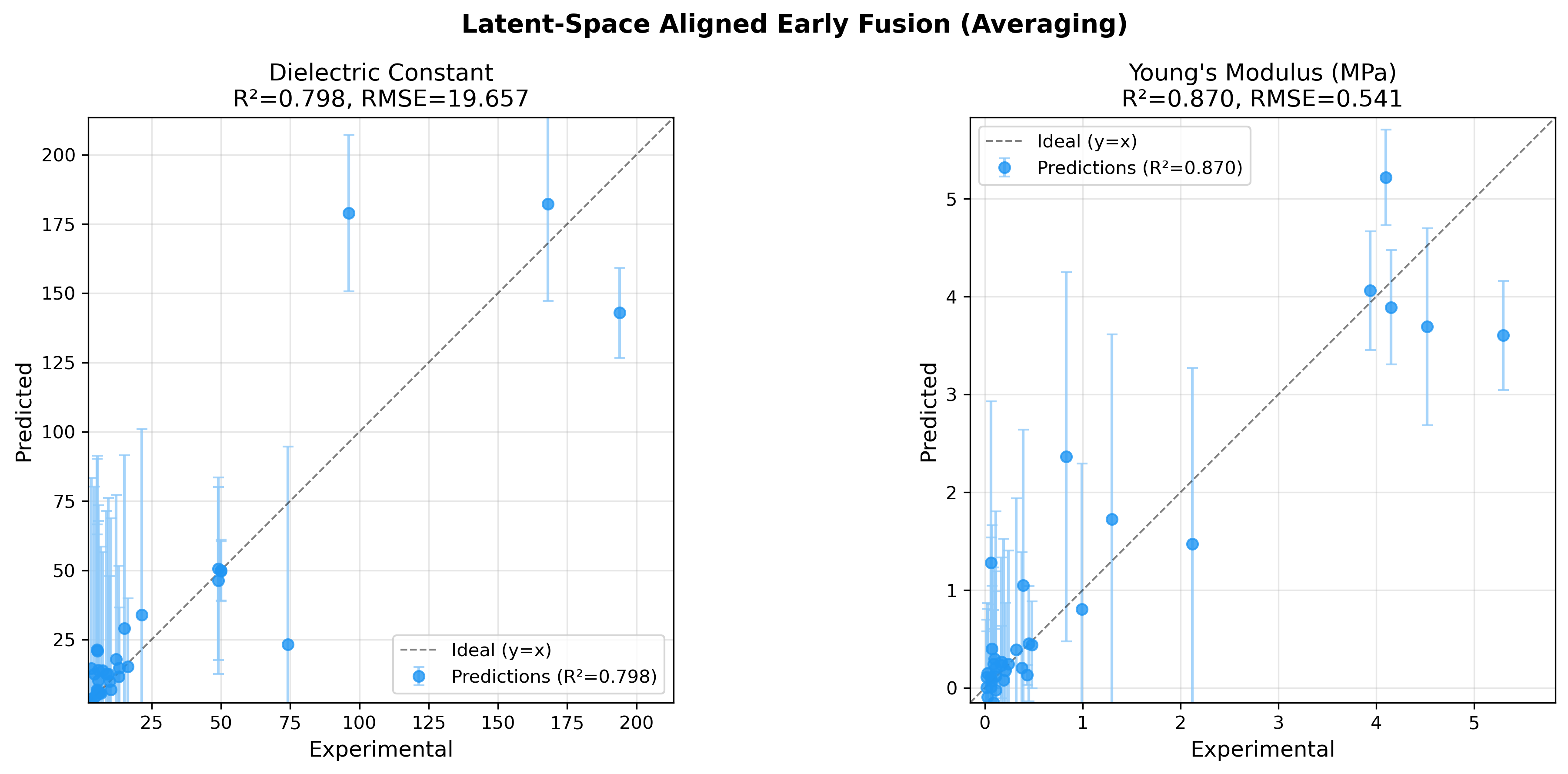}
\caption{Parity plots for dielectric constant ($k$) and Young’s modulus ($E$) using latent-space aligned early fusion (averaging). Error bars denote predictive uncertainty from GPR.}
\label{fig:aligned_parity}
\end{figure}

As shown in Table~\ref{tab:unimodal_vs_multimodal}, pretrained representations consistently outperform traditional descriptors under extreme data scarcity. Among unimodal models, TransPolymer achieves the strongest performance (mean $R^2 = 0.732$), followed by the pretrained GIN encoder (0.716) and PolyBERT (0.658), whereas Morgan fingerprints yield substantially lower predictive accuracy (0.542). These results highlight the advantage of pretrained polymer representations in low-data regimes. Integrating sequence and graph embeddings further improves predictive performance, achieving a mean $R^2$ of 0.834 and the lowest mean RMSE of 10.099, suggesting that the two modalities capture complementary structural and chemical information. Table~\ref{tab:fusion_strategies} further demonstrates that fusion strategy influences multimodal effectiveness. Naive early fusion yields moderate performance, with mean $R^2$ values of 0.733 (concatenation) and 0.735 (averaging), while prediction-level late fusion improves results to a mean $R^2$ of 0.791. The best overall performance among evaluated strategies is obtained using latent-space aligned early fusion with averaging (mean $R^2 = 0.834$). Although the dataset size limits formal statistical power, the performance gains are consistent across LOOCV folds and across both target properties, indicating robust cross-modal integration under extreme data scarcity. To visually assess predictive behavior, Figure~\ref{fig:aligned_parity} presents parity plots for dielectric constant and Young’s modulus. The predictions closely follow the ideal $y=x$ trend for both properties, demonstrating stable agreement between experimental and predicted values.

\begin{table}[ht]
\centering
\caption{Evaluation of unimodal and multimodal representations for elastomer property prediction.}
\label{tab:unimodal_vs_multimodal}
\resizebox{\columnwidth}{!}{
\begin{tabular}{llcccccc}
\toprule
 &  & \multicolumn{3}{c}{$R^2 \uparrow$} & \multicolumn{3}{c}{RMSE $\downarrow$} \\
\cmidrule(r){3-5} \cmidrule(r){6-8}
\textbf{Modality} & \textbf{Feature Representation} &
$k$ & $E$ & Mean &
$k$ & $E$ & Mean \\
\midrule

\multirow{3}{*}{Sequence}
    & Morgan Fingerprint 
    & $0.367 \pm 0.043$ & $0.716 \pm 0.025$ & $0.542 \pm 0.026$ 
    & $33.837 \pm 7.689$ & $0.766 \pm 0.211$ & $17.302 \pm 3.905$ \\
    
    & PolyBERT (Pretrained)
    & $0.492 \pm 0.030$ & $0.825 \pm 0.019$ & $0.658 \pm 0.017$ 
    & $30.101 \pm 7.820$ & $0.595 \pm 0.198$ & $15.348 \pm 3.966$ \\

    & TransPolymer (Pretrained) 
    & $0.628 \pm 0.034$ & $0.836 \pm 0.010$ & $0.732 \pm 0.018$ 
    & $26.113 \pm 4.934$ & $0.598 \pm 0.086$ & $13.356 \pm 2.473$ \\

\midrule

Graph 
    & GIN Encoder (Pretrained) 
    & $0.554 \pm 0.037$ & $\boldsymbol{0.877} \pm 0.009$ & $0.716 \pm 0.019$ 
    & $28.306 \pm 6.713$ & $\boldsymbol{0.517} \pm 0.070$ & $14.412 \pm 3.359$ \\

\midrule

Multimodal 
    & Ours 
    & $\boldsymbol{0.798} \pm 0.137$ 
    & $0.870 \pm 0.089$ 
    & $\boldsymbol{0.834} \pm 0.084$ 
    & $\boldsymbol{19.657} \pm 5.088$ 
    & $0.541 \pm 0.144$ 
    & $\boldsymbol{10.099} \pm 2.549$ \\

\bottomrule
\end{tabular}}
\end{table}

\begin{table}[ht]
\centering
\caption{Evaluation of multimodal fusion strategies for elastomer property prediction.}
\label{tab:fusion_strategies}
\resizebox{\columnwidth}{!}{
\begin{tabular}{llcccccc}
\toprule
 &  & \multicolumn{3}{c}{$R^2 \uparrow$} & \multicolumn{3}{c}{RMSE $\downarrow$} \\
\cmidrule(r){3-5} \cmidrule(r){6-8}
\textbf{Fusion Type} & \textbf{Method} &
$k$ & $E$ & Mean &
$k$ & $E$ & Mean \\
\midrule

\multirow{2}{*}{Early Fusion} 
    & Concatenation 
    & $0.654 \pm 0.056$ & $0.812 \pm 0.043$ & $0.733 \pm 0.031$ 
    & $25.666 \pm 2.152$ & $0.645 \pm 0.075$ & $13.155 \pm 1.070$ \\

    & Averaging     
    & $0.645 \pm 0.060$ & $0.824 \pm 0.022$ & $0.735 \pm 0.026$ 
    & $25.967 \pm 2.322$ & $0.627 \pm 0.038$ & $13.297 \pm 1.151$ \\

\midrule

\multirow{2}{*}{
\begin{tabular}{@{}l@{}}
Latent-Space Aligned\\
Early Fusion
\end{tabular}
}
    & Concatenation 
    & $0.638 \pm 0.134$ & $0.861 \pm 0.044$ & $0.749 \pm 0.061$ 
    & $25.916 \pm 4.676$ & $0.553 \pm 0.081$ & $13.234 \pm 2.319$ \\

    & Averaging     
    & $\boldsymbol{0.798} \pm 0.137$ 
    & $\boldsymbol{0.870} \pm 0.089$ 
    & $\boldsymbol{0.834} \pm 0.084$ 
    & $\boldsymbol{19.657} \pm 5.088$ 
    & $\boldsymbol{0.541} \pm 0.144$ 
    & $\boldsymbol{10.099} \pm 2.549$ \\
\midrule

Late Fusion 
    & Weighted Combination (Aligned, $\alpha=0.7$) 
    & $0.741 \pm 0.064$ 
    & $0.840 \pm 0.069$ 
    & $0.791 \pm 0.043$ 
    & $22.097 \pm 2.676$ 
    & $0.585 \pm 0.127$ 
    & $11.341 \pm 1.331$ \\

\bottomrule
\end{tabular}}
\end{table}

In this work, we demonstrate that pretrained multimodal polymer representations enable reliable prediction of dielectric constant and Young’s modulus under extreme data scarcity. By curating a standardized dataset of acrylate-based dielectric elastomers and integrating pretrained sequence-based and graph-based encoders, we show that multimodal learning consistently outperforms unimodal baselines. Among the evaluated strategies, latent-space aligned early fusion achieves the strongest overall performance, highlighting the importance of explicit cross-modal representation alignment for effective information integration in low-data regimes. Beyond the specific elastomer system studied here, our findings illustrate how pretrained multimodal polymer representations can be systematically transferred to small, specialized materials datasets. This data-efficient framework provides a practical pathway for leveraging large polymer corpora to support predictive modeling and the accelerated design of soft high-$k$ dielectric elastomers and related polymer systems under extreme data scarcity.

\section*{Data and Software Availability}

The curated dataset and all source code used in this study will be publicly available in our GitHub repository at \url{https://github.com/HySonLab/Polymers}.

\bibliography{main}

@article{amamoto2022data,
  title={Data-driven approaches for structure-property relationships in polymer science for prediction and understanding},
  author={Amamoto, Yoshifumi},
  journal={Polymer Journal},
  volume={54},
  number={8},
  pages={957--967},
  year={2022},
  publisher={Nature Publishing Group UK London}
}

@article{doan2020machine,
  title={Machine-learning predictions of polymer properties with Polymer Genome},
  author={Doan Tran, Huan and Kim, Chiho and Chen, Lihua and Chandrasekaran, Anand and Batra, Rohit and Venkatram, Shruti and Kamal, Deepak and Lightstone, Jordan P and Gurnani, Rishi and Shetty, Pranav and others},
  journal={Journal of Applied Physics},
  volume={128},
  number={17},
  year={2020},
  publisher={AIP Publishing}
}

@article{xu2023transpolymer,
  title={TransPolymer: a Transformer-based language model for polymer property predictions},
  author={Xu, Changwen and Wang, Yuyang and Barati Farimani, Amir},
  journal={npj Computational Materials},
  volume={9},
  number={1},
  pages={64},
  year={2023},
  publisher={Nature Publishing Group UK London}
}

@article{kuenneth2023polybert,
  title={polyBERT: a chemical language model to enable fully machine-driven ultrafast polymer informatics},
  author={Kuenneth, Christopher and Ramprasad, Rampi},
  journal={Nature communications},
  volume={14},
  number={1},
  pages={4099},
  year={2023},
  publisher={Nature Publishing Group UK London}
}

@article{ma2020pi1m,
  title={PI1M: a benchmark database for polymer informatics},
  author={Ma, Ruimin and Luo, Tengfei},
  journal={Journal of Chemical Information and Modeling},
  volume={60},
  number={10},
  pages={4684--4690},
  year={2020},
  publisher={ACS Publications}
}

@inproceedings{radford2021learning,
  title={Learning transferable visual models from natural language supervision},
  author={Radford, Alec and Kim, Jong Wook and Hallacy, Chris and Ramesh, Aditya and Goh, Gabriel and Agarwal, Sandhini and Sastry, Girish and Askell, Amanda and Mishkin, Pamela and Clark, Jack and others},
  booktitle={International conference on machine learning},
  pages={8748--8763},
  year={2021},
  organization={PmLR}
}

@article{Feng2024NatCommun,
  title   = {A large-strain and ultrahigh energy density dielectric elastomer for fast moving soft robot},
  author  = {Feng, Wenwen and Sun, Lin and Jin, Zhekai and Chen, Lili and Liu, Yuncong and Xu, Hao and Wang, Chao},
  journal = {Nature Communications},
  year    = {2024},
  volume  = {15},
  pages   = {4222},
  doi     = {10.1038/s41467-024-48243-y}
}

@article{Yin2024NatCommun54278,
  title   = {A high-response-frequency bimodal network polyacrylate elastomer with ultrahigh power density under low electric field},
  author  = {Yin, Li-Juan and Du, Boyuan and Hu, Hui-Yi and Dong, Wen-Zhuo and Zhao, Yu and Zhang, Zili and Zhao, Huichan and Zhong, Shao-Long and Yi, Chenyi and Qu, Liangti and Dang, Zhi-Min},
  journal = {Nature Communications},
  year    = {2024},
  volume  = {15},
  number  = {9819},
  pages   = {9819},
  doi     = {10.1038/s41467-024-54278-y},
  url     = {https://www.nature.com/articles/s41467-024-54278-y},
  month   = nov
}

@article{Han2023HybridPrestrainLocked,
  title   = {Hybrid Fabrication of Prestrain-Locked Acrylic Dielectric Elastomer Thin Films and Multilayer Stacks},
  author  = {Han, Ziqing and Peng, Zihang and Guo, Yuxuan and Wang, Huiying and Plamthottam, Roshan and Pei, Qibing},
  journal = {Macromolecular Rapid Communications},
  year    = {2023},
  volume  = {44},
  number  = {15},
  pages   = {e2300160},
  doi     = {10.1002/marc.202300160}
}

@article{Shi2022ProcessablePHDE,
  title   = {A processable, high-performance dielectric elastomer and multilayering process},
  author  = {Shi, Ye and Askounis, Erin and Plamthottam, Roshan and Libby, Tom and Peng, Zihang and Youssef, Kareem and Pu, Junhong and Pelrine, Ron and Pei, Qibing},
  journal = {Science},
  year    = {2022},
  volume  = {377},
  number  = {6602},
  pages   = {228--232},
  doi     = {10.1126/science.abn0099},
  url     = {https://doi.org/10.1126/science.abn0099}
}

@article{Zhao2022AdvancedAcrylateDE,
  title   = {Advanced Acrylate Dielectric Elastomers with Large Actuation Strains at Very Low Electric Field},
  author  = {Zhao, Yu and Feng, Qi-Yao and Xie, Ya-Kun and Zhang, Zi-Li and Yin, Li-Juan and Dang, Zhi-Min},
  journal = {ACS Applied Polymer Materials},
  year    = {2022},
  volume  = {4},
  number  = {12},
  pages   = {8892--8899},
  doi     = {10.1021/acsapm.2c01293},
  url     = {https://doi.org/10.1021/acsapm.2c01293}
}

@article{Xiong2023TailoringCrosslinkingNetworks,
  title   = {Tailoring crosslinking networks to fabricate photocurable polyurethane acrylate (PUA) dielectric elastomer with balanced electromechanical performance},
  author  = {Xiong, Lulu and Li, Delong and Yang, Yongfei and Ye, Xiaoxiao and Huang, Yu and Xu, E. and Xia, Chuanhui and Yang, Mingbo and Liu, Zhengying and Cui, Xudong and Wang, Feng and Huang, Yanhao},
  journal = {Reactive and Functional Polymers},
  year    = {2023},
  volume  = {183},
  pages   = {105498},
  month   = feb,
  doi     = {10.1016/j.reactfunctpolym.2023.105498},
  url     = {https://www.sciencedirect.com/science/article/pii/S1381514823000019}
}

@inproceedings{Ha2008HighElectromechanicalIPN,
  title     = {High electromechanical performance of electroelastomers based on interpenetrating polymer networks},
  author    = {Ha, Soon Mok and Park, Il Seok and Wissler, Michael and Pelrine, Ron and Stanford, Scott and Kim, Kwang J. and Kovacs, Gabor and Pei, Qibing},
  booktitle = {Electroactive Polymer Actuators and Devices (EAPAD) 2008},
  editor    = {Bar-Cohen, Yoseph},
  series    = {Proceedings of SPIE},
  volume    = {6927},
  pages     = {69272C},
  year      = {2008},
  publisher = {SPIE},
  address   = {Bellingham, WA},
  doi       = {10.1117/12.778282},
  url       = {https://doi.org/10.1117/12.778282},
  note      = {Held 10--13 March 2008, San Diego, California, USA}
}

@article{Liu2022TernaryDipolarDE,
  title   = {Significant improvements in the electromechanical performance of dielectric elastomers by introducing ternary dipolar groups},
  author  = {Liu, Leipeng and Zhang, Kangning and Liu, Jinru and Zhu, Lei and Xie, Ruiying and Lv, Shenghua},
  journal = {Reactive and Functional Polymers},
  year    = {2022},
  volume  = {172},
  pages   = {105177},
  doi     = {10.1016/j.reactfunctpolym.2022.105177},
  url     = {https://doi.org/10.1016/j.reactfunctpolym.2022.105177}
}

@article{Tan2019PolarCrosslinkingDEA,
  title   = {Enhancing dynamic actuation performance of dielectric elastomer actuators by tuning viscoelastic effects with polar crosslinking},
  author  = {Tan, Matthew Wei Ming and Thangavel, Gurunathan and Lee, Pooi See},
  journal = {NPG Asia Materials},
  year    = {2019},
  volume  = {11},
  number  = {1},
  pages   = {62},
  doi     = {10.1038/s41427-019-0147-5},
  url     = {https://www.nature.com/articles/s41427-019-0147-5},
  month   = oct,
  note    = {Published: 25 Oct 2019}
}

@article{Shao2019AEgOANI,
  title   = {A novel high dielectric constant acrylic resin elastomer nanocomposite with pendant oligoanilines},
  author  = {Shao, Jing and Wang, Jing-Wen and Wei, Lei and Wu, Sen-Qiang and Yang, Yan-Hua and Ren, Hua},
  journal = {Composites Part B: Engineering},
  year    = {2019},
  volume  = {176},
  pages   = {107216},
  doi     = {10.1016/j.compositesb.2019.107216},
  url     = {https://doi.org/10.1016/j.compositesb.2019.107216},
  note    = {Available online 23 July 2019}
}

@article{Gao2021ImprovingDielectricAE_PS_RGO,
  title   = {Improving the dielectric properties of acrylic resin elastomer with reduced graphene oxide decorated with polystyrene},
  author  = {Gao, Xin-Hua and Wang, Jing-Wen and Liu, Da-Nian and Wang, Xin-Zhu and Wang, Hou-Qing and Wei, Lei and Ren, Hua},
  journal = {European Polymer Journal},
  year    = {2021},
  volume  = {150},
  pages   = {110418},
  doi     = {10.1016/j.eurpolymj.2021.110418},
  url     = {https://doi.org/10.1016/j.eurpolymj.2021.110418}
}

@article{Han2023HybridFabrication,
  title   = {Hybrid Fabrication of Prestrain-Locked Acrylic Dielectric Elastomer Thin Films and Multilayer Stacks},
  author  = {Han, Ziqing and Peng, Zihang and Guo, Yuxuan and Wang, Huiying and Plamthottam, Roshan and Pei, Qibing},
  journal = {Macromolecular Rapid Communications},
  year    = {2023},
  volume  = {44},
  number  = {15},
  pages   = {e2300160},
  doi     = {10.1002/marc.202300160}
}

@article{Zhao2018Remarkable,
  title   = {Remarkable electrically actuation performance in advanced acrylic-based dielectric elastomers without pre-strain at very low driving electric field},
  author  = {Zhao, Yu and others},
  journal = {Polymer},
  year    = {2018},
  volume  = {137},
  pages   = {269--275},
  doi     = {10.1016/j.polymer.2017.12.065}
}

@article{CEJ2021UltrasoftPentablock,
  title   = {Ultrasoft-yet-strong pentablock copolymer for dielectric elastomer},
author={Zheqi Chen},
  journal = {Chemical Engineering Journal},
  year    = {2021},
  volume  = {405},
  pages   = {126634},
  doi     = {10.1016/j.cej.2020.126634}
}

@article{Sung2025HighkZwitterionic,
  title   = {High-k zwitterionic dielectric elastomers with internal plasticization for low-voltage actuation},
  author  = {Sung, Gimin and Yu, Changhoon and Park, Jae-Man and Lee, Yun Hyeok and Park, Chang Seo and Lee, Hakjun and Kwon, Min Sang and Sun, Jeong-Yun},
  journal = {Materials Today},
  year    = {2025},
  volume  = {88},
  month   = sep,
  pages   = {109--116},
  doi     = {10.1016/j.mattod.2025.05.023},
}

@article{Ankit2020HighkUltrastretchable,
  title   = {High-k, Ultrastretchable Self-Enclosed Ionic Liquid-Elastomer Composites for Soft Robotics and Flexible Electronics},
  author  = {{Ankit} and Tiwari, Naveen and Ho, Fanny and Krisnadi, Febby and Kulkarni, Mohit Rameshchandra and Nguyen, Linh Lan and Koh, Soo Jin Adrian and Mathews, Nripan},
  journal = {ACS Applied Materials \& Interfaces},
  year    = {2020},
  volume  = {12},
  number  = {33},
  pages   = {37561--37570},
  month   = aug,
  doi     = {10.1021/acsami.0c08754},
}

@misc{Li2024SSRNPolarCyano,
  title        = {Significant Improvements in the Dielectric Performance of Dielectric Elastomers with Polar Cyano Groups},
  author       = {Li, Ying and Zhao, Qiyun and Huang, Jingjing and Hu, Xiaoqian and Yan, Yuanhu and Li, Lu},
  year         = {2024},
  month        = nov,
  note         = {SSRN working paper. Posted: 2024-11-23},
  doi          = {10.2139/ssrn.5031777},
  howpublished = {SSRN},
  url          = {https://ssrn.com/abstract=5031777},
}

@article{Wang2023PILFillerDEA,
  title   = {A highly stretchable, self-healable, transparent and solid-state poly(ionic liquid) filler for high-performance dielectric elastomer actuators},
  author  = {Wang, Hui and Tan, Matthew Wei Ming and Poh, Wei Church and Gao, Dace and Wu, Wenting and Lee, Pooi See},
  journal = {Journal of Materials Chemistry A},
  year    = {2023},
  volume  = {11},
  number  = {26},
  pages   = {14159--14168},
  doi     = {10.1039/D3TA01954C},
}

@article{Adeli2024BottlebrushDEG,
  title   = {Synthesis of Bottlebrush Polymers with Spontaneous Self-Assembly for Dielectric Generators},
  author  = {Adeli, Yeerlan and Venkatesan, Thulasinath Raman and Mezzenga, Raffaele and N{\"u}esch, Frank A and Opris, Dorina M},
  journal = {ACS Applied Polymer Materials},
  year    = {2024},
  volume  = {6},
  number  = {9},
  pages   = {4999--5010},
  month   = apr,
  doi     = {10.1021/acsapm.3c03053},
}

@article{Park2024GlassTransitionSelfHeal,
  title   = {Glass transition temperature as a unified parameter to design self-healable elastomers},
  author  = {Park, Jae-Man and Park, Chang Seo and Kwak, Sang Kyu and Sun, Jeong-Yun},
  journal = {Science Advances},
  year    = {2024},
  volume  = {10},
  number  = {28},
  pages   = {eadp0729},
  month   = jul,
  doi     = {10.1126/sciadv.adp0729},
}

@article{Zhang2025DEASoftRoboticsReview,
  title   = {A Review of the Applications and Challenges of Dielectric Elastomer Actuators in Soft Robotics},
  author  = {Zhang, Qinghai and Yu, Wei and Zhao, Jianghua and Meng, Chuizhou and Guo, Shijie},
  journal = {Machines},
  year    = {2025},
  volume  = {13},
  number  = {2},
  pages   = {101},
  doi     = {10.3390/machines13020101},
  url     = {https://www.mdpi.com/2075-1702/13/2/101}
}

@article{Yin2021SoftToughFastPolyacrylate,
  title   = {Soft, tough, and fast polyacrylate dielectric elastomer for non-magnetic motor},
  author  = {Yin, Li-Juan and Zhao, Yu and Zhu, Jing and Yang, Minhao and Zhao, Huichan and Pei, Jia-Yao and Zhong, Shao-Long and Dang, Zhi-Min},
  journal = {Nature Communications},
  year    = {2021},
  volume  = {12},
  pages   = {4517},
  doi     = {10.1038/s41467-021-24851-w},
  url     = {https://www.nature.com/articles/s41467-021-24851-w}
}

@article{Liu2022TernaryDipolarGroups,
  title   = {Significant improvements in the electromechanical performance of dielectric elastomers by introducing ternary dipolar groups},
  author  = {Liu, Leipeng and Zhang, Kangning and Liu, Jinru and Zhu, Lei and Xie, Ruiying and Lv, Shenghua},
  journal = {Reactive and Functional Polymers},
  year    = {2022},
  volume  = {172},
  month   = mar,
  pages   = {105177},
  doi     = {10.1016/j.reactfunctpolym.2022.105177},
  url     = {https://www.sciencedirect.com/science/article/pii/S1381514822000190}
}

@article{Shi2018DielectricGels,
  title   = {Dielectric gels with ultra-high dielectric constant, low elastic modulus, and excellent transparency},
  author  = {Shi, Lei and Yang, Ruisen and Lu, Shiyao and Jia, Kun and Xiao, Chunhui and Lu, Tongqing and Wang, Tiejun and Wei, Wei and Tan, Hui and Ding, Shujiang},
  journal = {NPG Asia Materials},
  year    = {2018},
  volume  = {10},
  pages   = {821--826},
  doi     = {10.1038/s41427-018-0077-7},
  url     = {https://www.nature.com/articles/s41427-018-0077-7}
}

@article{Bezsudnov2023DEAMaterialsDesign,
  title     = {Dielectric elastomer actuators: materials and design},
  author    = {Bezsudnov, Igor V. and Khmelnitskaia, Alina G. and Kalinina, Aleksandra A. and Ponomarenko, Sergey A.},
  journal   = {Russian Chemical Reviews},
  year      = {2023},
  volume    = {92},
  number    = {2},
  pages     = {RCR5070},
  doi       = {10.57634/RCR5070},
  url       = {https://rcr.colab.ws/publications/10.57634/RCR5070}
}

\end{document}